\documentclass[11pt]{article}
\usepackage[utf8]{inputenc}
\usepackage[a4paper,margin=1in]{geometry}

\usepackage{float}
\usepackage{amsmath}
\usepackage{amsfonts}
\usepackage{amsthm}
\usepackage{hyperref}

\usepackage{amssymb}
 \usepackage{graphicx}
 \usepackage{titling}
\usepackage[
  backend=bibtex,        
  style=authoryear,
  maxcitenames=2,
  maxbibnames=99,
  uniquename=false,
  dashed=false,
  url=true, doi=true, eprint=true
]{biblatex}

\newtheorem{theorem}{Theorem}[section]

\theoremstyle{definition}
\newtheorem{definition}[theorem]{Definition}

\theoremstyle{remark}

 \title{Using physics-inspired Singular Learning Theory to understand grokking \& other phase transitions in modern neural networks}
\author{%
  Anish Lakkapragada\\
  Department of Statistics \\ Yale University\\[0.2em]
  \texttt{anish.lakkapragada@yale.edu}
}
 
 \usepackage{fancyhdr}
\fancypagestyle{plain}{
    \fancyhf{} 
}
\makeatletter
\def\@maketitle{%
  \newpage
  \begin{center}%
    \let\footnote\thanks
    {\LARGE \@title\par}%
    \vskip 1em
    {\large \@author\par}
    \vskip 0.5em
  \end{center}%
  \par}
\makeatother

\usepackage{lipsum}  
\usepackage{cmbright}

\makeatletter

\DeclareCiteCommand{\parencite}[\mkbibparens]
  {\usebibmacro{prenote}}
  {\usebibmacro{citeindex}%
   \printtext[bibhyperref]{\printnames{labelname}\addcomma}
   \addnbspace
   \printtext[bibhyperref]{\printlabeldateextra}}          
  {\multicitedelim\addsemicolon\space}
  {\usebibmacro{postnote}}

\DeclareCiteCommand{\textcite}
  {\usebibmacro{prenote}}
  {\usebibmacro{citeindex}%
   \printtext[bibhyperref]{\printnames{labelname}}         
   \addnbspace
   \printtext[bibhyperref]{\mkbibparens{\printlabeldateextra}}} 
  {\multicitedelim\addsemicolon\space}
  {\usebibmacro{postnote}}
\makeatother

\begin{document}

\maketitle

\begin{abstract}
Classical statistical inference and learning theory often fail to explain the success of modern neural networks. A key reason is that these models are non-identifiable (singular), violating core assumptions behind PAC bounds and asymptotic normality. Singular learning theory (SLT), a physics-inspired framework grounded in algebraic geometry, has gained popularity for its ability to close this theory–practice gap. In this paper, we empirically study SLT in toy settings relevant to interpretability and phase transitions. First, we understand the SLT free energy $\mathcal{F}_n$ by testing an Arrhenius-style rate hypothesis using both a grokking modulo-arithmetic model and Anthropic’s Toy Models of Superposition. Second, we understand the local learning coefficient $\lambda_{\alpha}$ by measuring how it scales with problem difficulty across several controlled network families (polynomial regressors, low-rank linear networks, and low-rank autoencoders). Our experiments recover known scaling laws while others yield meaningful deviations from theoretical expectations. Overall, our paper illustrates the many merits of SLT for understanding neural network phase transitions, and poses open research questions for the field.
\end{abstract}

\section{Introduction}

Classical statistics are insufficient to understand modern machine-learning. For example, well-studied PAC-bounds from statistical learning theory are vacuous and fail to explain the remarkable generalization power of neural networks. The reason for this comes from a well-kept secret for the last 15 years: neural networks are \emph{singular} statistical models \parencite{wei2022deep}. Namely, unlike \emph{regular} models, singular models can implement the same function through multiple distinct parameter values. As an example of how this could occur in a singular model, consider that $\forall \alpha > 0, \text{ReLU}(x) = \frac{1}{\alpha} \text{ReLU}(\alpha x)$ \parencite{hoogland2023weirdtrick}. Moreover, singular models do \emph{not} exhibit expected statistical behavior. As an example, the Fisher Information matrix, the basis of asymptotic normality theory for MLEs, is often non-invertible at the true parameters in singular models \parencite{watanabe2010equations}. Most pressingly, the majority of AI models today (neural networks, LLMs) are singular. Thus, out of a dire need to study them for AI safety \parencite{lehalleur2025you}, the extremely mathematically rigorous field of singular learning theory (SLT) developed in 2009 has gained renewed attention. 


\subsection*{Introduction to Singular Learning Theory (SLT)}

SLT is an extremely mathematically rigorous framework developed from algebraic geometry by Dr. Sumio Watanabe in his two books \parencite{watanabe2009} and \parencite{watanabe2018}. At its core, SLT posits that the \emph{singularities} of the model -- the weights at which the model is non-identifiable\footnote{The weights $w \in \mathcal{W}$ are non-identifiable if $\exists \ w' \in \mathcal{W}$ s.t. the model functions $f(x, w) = f(x, w')$. } --- determine the weight spaces that the model will occupy as $n \to \infty$. More concretely, if we assume our model has parameter space $\mathcal{W} \subseteq \mathbb{R}^d$ and we take $\{\mathcal{W}_{\alpha}\}_{\alpha}$ to be a sequence of subsets of $\mathcal{W}$ satisfying mild analytical conditions\footnote{For a full discussion of these properties, please see Section 4.1 in \parencite{chen2023dynamical}.}, then through application of \parencite[\S~6.3]{watanabe2018} we get the following approximation: 

\[
\mathcal{F}_n := - \log p(D_n)  \approx \min_{\alpha} [nL_n(w^*_{\alpha}) + \lambda_\alpha \log n ] 
\]

where $\mathcal{F}_n$ is the ``free energy" of the model, $w_\alpha^*$ is a $\mathcal{W}_{\alpha}$-global minima of the population negative-log-likelihood, $L_n$ is the empirical negative-log-likelihood, and $\lambda_\alpha$ is the local learning coefficient (LLC\footnote{Note that higher values of $\lambda_{\alpha}$ indicate greater model complexity.}) \parencite{lau2023local}. From this equation, we gain a principled understanding of the \emph{internal model selection} a neural network is performing during training \parencite{chen2023dynamical}. Specifically, we see that a neural network during training is consistently trying to minimize the free-energy through a tradeoff of the loss $L_n(w^*_{\alpha})$ and model complexity $\lambda_{\alpha}$.

\subsection*{Research Questions}

We are interested in the following broad question: \emph{how can we understand\footnote{Understand in the when/what/where/why setting.} grokking and other phase transitions in neural networks through singular learning theory?} Two main research questions we tackle in this vein are: 

\begin{itemize}
    \item How far can we extend SLT's physics-based ``free-energy" quantity $\mathcal{F}_n$ to understand \emph{when} a model will undergo a phase transition or grok?
    \item Can we better understand SLT's local learning coefficient $\lambda_{\alpha}$ \parencite{lau2023local} and its sensitivity to various problems and their difficulty?
\end{itemize}

\section{Methods}

We aim to use lightweight models known to grok in our study of SLT, as we can train them many times to gain good statistics. 

\paragraph{LLC Estimation.} We are interested in the free energy of these models over time, which is fundamentally based on SLT's local learning coefficient $\lambda_{\alpha}$. Measuring the $\lambda_{\alpha}$ of a model is extremely non-trivial, and so we will defer to using \parencite{lau2023local}'s Stochastic Gradient Langevin Dynamics (SGLD) MCMC estimation procedure. All SGLD hyperparameters were tuned for stability and are summarized in the Appendix.



\section{Experiments}

\subsection*{Question 1: Understanding the temporality of grokking and phase transitions}

\paragraph{Our Arrhenius Reaction Rate Hypothesis.} Suppose we are training a model, and let us define $\mathcal{W}_t$ to be the weights of the model at iteration $t$. Now suppose at iteration $i$ the model has only memorized (i.e. 100\% training accuracy but poor testing accuracy) whereas at iteration $j$ the model has grokked (i.e. 100\% testing \& training accuracy). Let us define $\mathcal{F}_i$ and $\mathcal{F}_j$ to be the model's free energy, as per the SLT definition, at each of these respective iterations. Borrowing from the Arrhenius Reaction Rate Theory in chemical kinetics \parencite{arrhenius1889reaktionsgeschwindigkeit}, we present the following hypothesis to explain the time it takes to grok: 

\[
\boxed{r_{i\to j} \propto \exp\Big( \,\beta_{\text{eff}}\, \Delta \mathcal{F}_{i\to j} \Big), \quad r_{i \to j} := j - i, \quad \Delta \mathcal{F}_{i \to j} := \mathcal{F}_i - \mathcal{F}_j}
\]

where $\Delta \mathcal{F}_{i \to j} < 0$ and $\beta_{\text{eff}}$ is an effective inverse temperature dependent on global hyperparameters (learning rate, batch size, etc.). The main intuition here is that the amount of time it takes for a phase transition to occur is exponential in how much it decreases the (free) energy of the model. 

\paragraph{Experiment One: Grokking on Modulo Arithmetic.} 

\begin{figure}
\begin{center}
\includegraphics[scale=0.3]{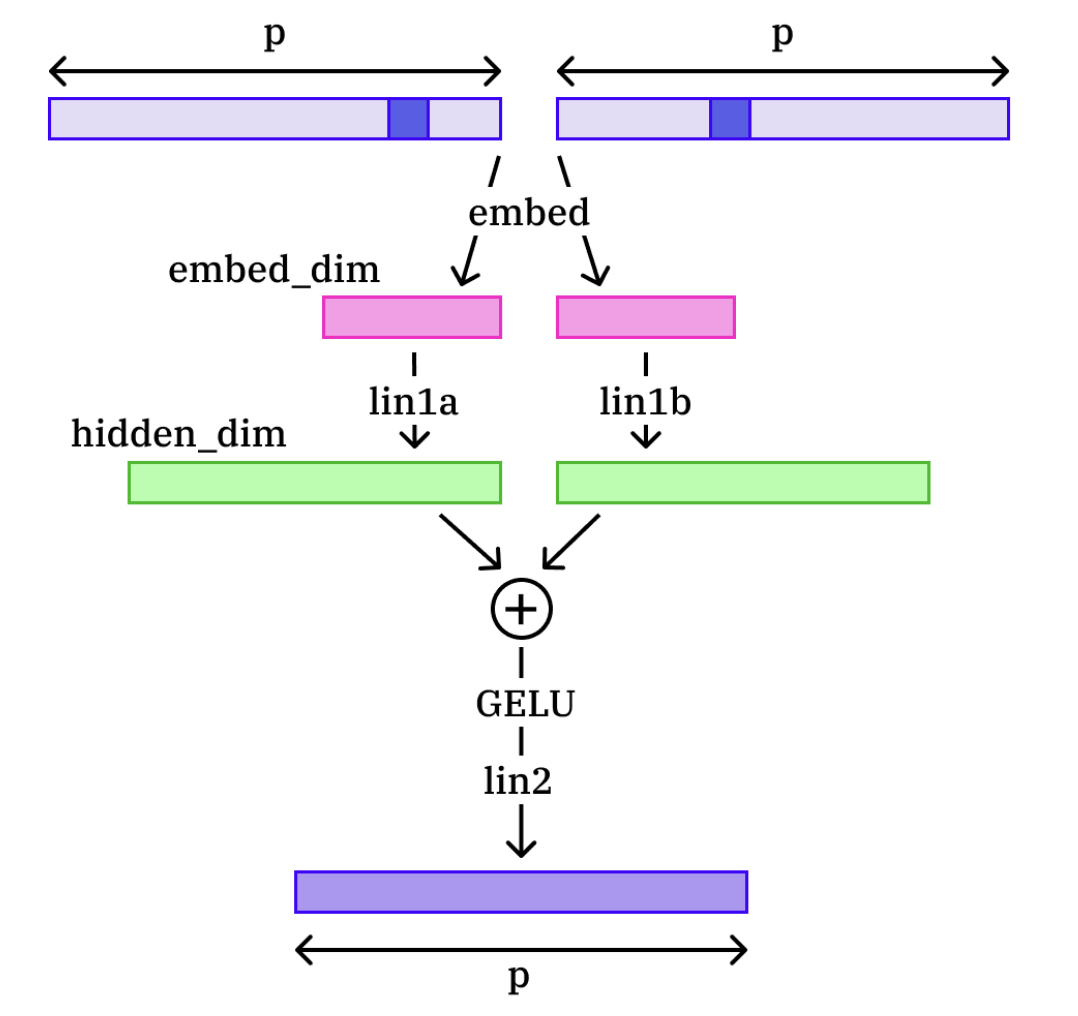}  
\end{center}
\caption{Diagram of the modulo arithmetic neural network we use for our experiments, taken from \parencite{panickssery2023learningcoef}. Note that this task can be thought of as taking two numbers in $\mathbb{Z}_p$ and outputting the modulo in $\mathbb{Z}_p$.}
\label{fig:modulo-arithmetic-arch}
\end{figure}

We first test our hypothesis on the setting of modulo arithmetic, which is a common setup to study grokking \parencite{power2022grokking, panickssery2023learningcoef, pearce2023memorize}. In more detail, we are training a neural network to $(a, b) \mapsto a + b \text{ mod } p$ for some prime $p$, using the architecture shown in Figure \ref{fig:modulo-arithmetic-arch}.  

For our specific experiments, we take $p = 53$ and train all our models using a randomly chosen $40\%$ subset of $\mathbb{Z}_{53} \times \mathbb{Z}_{53}$ (remainder $60\%$ was used for validation.) We trained 500 different models and for each of them stored\footnote{Time-based metrics were stored across 100 linearly-spaced checkpoints during training.} their training loss \& accuracy, validation loss \& accuracy, and if grokking occurred -- computed pre-grok and post-grok LLCs. We stored all of this information through use of a \href{https://wandb.ai}{Weights \& Biases} project; in total this process took over 7 days of continuous CPU compute to run. 

\begin{figure}[h!]
\centering
\includegraphics[width=0.7\linewidth]{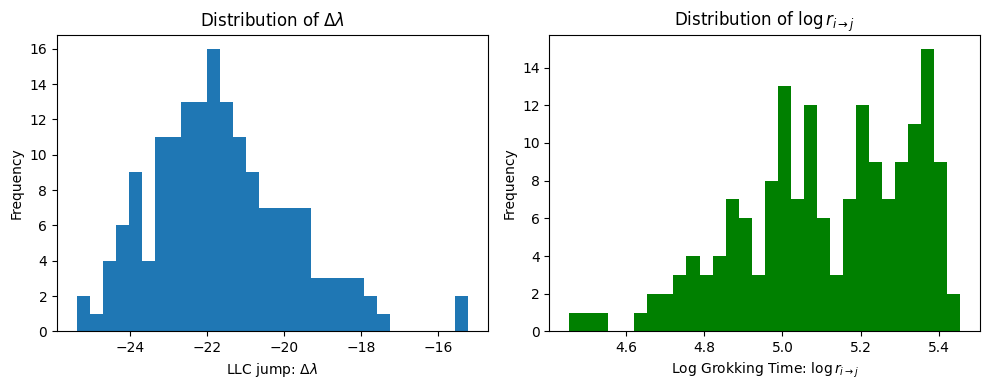}
\caption{Distribution of $\Delta\lambda$ and $\log r_{i \to j}$.}
\label{fig:distributions}
\end{figure}

\begin{figure}[h!]
\centering
\includegraphics[width=0.6\linewidth]{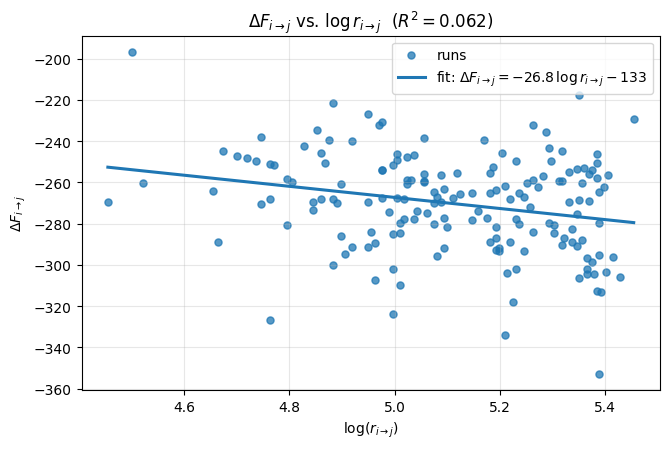}
\caption{$\Delta \mathcal{F}_{i \to j}$ vs.\ $\log r_{i \to j}$ with linear fit on modulo arithmetic toy networks with $p = 53$.}
\label{fig:free_energy_vs_time}
\end{figure}

For our analyses, which we present momentarily, we only considered the 168 (33.6\%) runs during which grokking occurred. We report a histogram of the distribution of observed LLC jumps $\Delta\lambda$ and log times for grokking $\log(r_{i \to j})$ in Figure \ref{fig:distributions}. We report a plot of $\log(r_{i \to j})$ versus $\Delta \mathcal{F}_{i \to j}$ across our runs in Figure \ref{fig:free_energy_vs_time}.

This observed relationship is \textbf{consistent with our hypothesis} as it is demonstrates a consistent negative slope, albeit with a low $R^2$ score. 

\paragraph{Experiment Two: Phase Transitions on Anthropic's Toy Models of Superposition.} To re-test our hypothesis on another problem, we perform the same experiment on Anthropic's Toy Models of Superposition (TMS) \parencite{elhage2022toy}. TMS are extremely small networks that reliably undergo aesthetic phase transitions, where the weight columns (when visualized) create increasingly complex shapes (e.g. $2$-gon $\to 3$-gon $\to 4-$gon). Note that this is in contrast to grokking, as now \emph{multiple} phase transitions can occur in a single run. 

\begin{figure}[h!]
\centering
\includegraphics[width=0.6\linewidth]{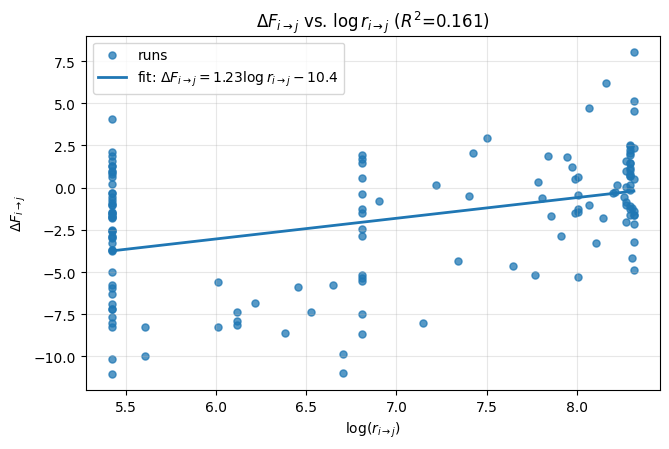}
\caption{$\Delta \mathcal{F}_{i \to j}$ vs.\ $\log r_{i \to j}$ with linear fit on Anthropic's TMS models.}
\label{fig:free_energy_vs_time_tms}
\end{figure}

This experiment is largely similar to the first, except for a few changes. We trained $60$ models (with Weights \& Biases again) for $4500$ iterations. For each run, we tracked the LLC across 100 linearly \& logarithmically-spaced checkpoints. Moreover, for a given run we detected transitions by (1) smoothing out the training loss curve across every $\sim 10$ steps and (2) finding segments in the training loss curve where the loss decreased by $\geq10\%$ of the total loss decrease throughout training. \emph{We observed our end results to vary greatly based on the method we used to detect these transitions.} We then computed $\Delta \mathcal{F}_{i \to j}$to be the change in free-energy across any two \emph{consecutive} transitions. We provide a plot of $\Delta \mathcal{F}_{i \to j}$ versus $\log r_{i \to j}$ with linear fit in Figure \ref{fig:free_energy_vs_time_tms}. 

While these results vary drastically from those in our prior experiment, we contend that we can still glean valuable information from them. We first address the three vertical clusters of transition times: we conjecture these clusters occurred as a result of (1) our method for detecting transitions and (2) uniqueness of the TMS experiment setup (i.e. an extremely low parameter model may have more fixed transition timesteps). Moreover, the linear fit demonstrated has an upward slope, which we note was highly sensitive to (1): we actually recovered a downward-sloping linear fit using a more rudimentary, non-smoothing transition detection approach. Given the variability in our results, we deemed this experiment as \textbf{inconclusive to our hypothesis} and proceeded to Question 2.

\subsection*{Question 2: Understanding how the LLC $\lambda_{\alpha}$ Scales with Problem Difficulty}

\paragraph{Motivation.} The second broad question we tackle in this project is understanding the behavior of the local learning coefficient (LLC) $\lambda_{\alpha}$. For context, the LLC is a complexity measure of a model ($\uparrow \text{LLC} \implies \uparrow \text{complexity}$) designed to be robust in measuring the complexity of singular models. However, the LLC is extremely unintuitive as seen by its definition (reproduced from \parencite{lau2023local}) below: 

\begin{definition}[Local Learning Coefficient (LLC)]
\emph{There exists a unique rational number $\lambda(w^*)$, a positive integer
$m(w^*)$ and some constant $c > 0$ such that, asymptotically as
$\varepsilon \to 0$,}
\[
  V(\varepsilon)
  = c \,\varepsilon^{\lambda(w^*)} (-\log \varepsilon)^{m(w^*)-1}
    + o\!\left( \varepsilon^{\lambda(w^*)} (-\log \varepsilon)^{m(w^*)-1} \right).
\]
\emph{We call $\lambda(w^*)$ the Local Learning Coefficient (LLC), and
$m(w^*)$ the local multiplicity.} 
\end{definition}

This is in stark contrast to \emph{statistical} learning theory's famous complexity measures (e.g. Vapnik–Chervonenkis Dimension), which are not only perhaps more intuitive but very well-studied. As such, we aim to take a stab at understanding the LLC by testing how well it scales with the problem difficulty. These experiments were inspired by \parencite{panickssery2023learningcoef}, who demonstrated that the LLC linearly scaled in (generalizing) modulo arithmetic networks with $p$. 

\paragraph{Experiment One: Polynomial Regressors of Increasing Degree.}

The first problem setup we study are univariate polynomial regressors of increasing degree. We say a function $f: \mathbb{R} \to \mathbb{R}$ is a polynomial regressor of degree $d$ if $f(x) = \sum_{i = 0}^d a_ix^i$ for $\{a_i\}_{i = 0}^d \subset \mathbb{R}$ and all $x_i \in \mathcal{X} \subseteq \mathbb{R}$. Note that the use of this constrained instance space $\mathcal{X}$ is required, as it is numerically unstable to use $x_i$ with absolute magnitude greater than one as we scale the degree $d \sim 10^3$. For a given degree $d$ and  instance space $\mathcal{X}$, we run the following procedure ten times to get an accurate estimate of the mean LLC $\lambda_d$ and its corresponding standard deviation: 

\begin{enumerate}
    \item Generate a $d$-degree dataset $\mathcal{D}_N = \{(x_i, y_i)\}_{i = 1}^N$ of $N = 500$ samples realizable by some polynomial regressor with all coefficients in $[-1, 1]$ and all samples $x_i \in \mathcal{X}$.
    
    \item Initialize a polynomial regressor of degree $d$ from scratch and train it on $\mathcal{D}_N$ until convergence.
    
    \item Measure the LLC of this converged model.
\end{enumerate}

\begin{figure}[h!]
\centering
\includegraphics[width=0.7\linewidth]{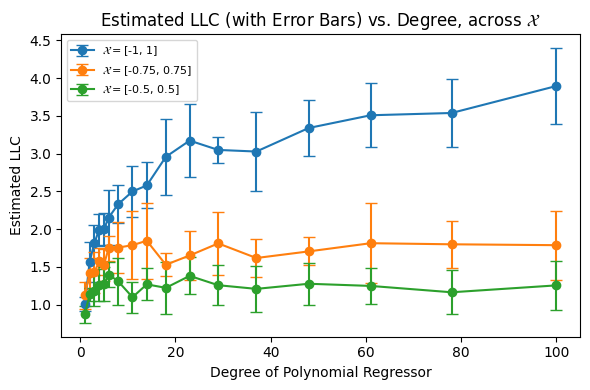}
\caption{Estimated LLC versus degree of polynomial regressor for all choices of instance space $\mathcal{X}$.}
\label{fig:llc_vs_poly_reg}
\end{figure}

We use $\mathcal{X} = [-1, 1], [-0.75, 0.75], $ and $[-0.5, 0.5]$ for our experiments. For each choice of $\mathcal{X}$, we present $\lambda_d$ for twenty choices of $d$ from $10^0$ to $10^3$ in Figure \ref{fig:llc_vs_poly_reg}.

While deceptively simple, this result is actually quite interesting. Observe that polynomial regressors are \emph{regular} models, meaning that there is a bijection from $\{a_i\}_{i = 0}^d$ to some function from $\mathbb{R} \to \mathbb{R}$. As such, following well-established results from SLT \parencite{wei2022deep} we would expect $\lambda_d = \frac{d}{2}$. This means our empirical LLC estimates, across all $\mathcal{X}$, are \emph{lower} than the theoretical expectations. We reason this to be the case because the input of our polynomial regressors is not $\mathbb{R}$ but instead some choice of $\mathcal{X} \subset \mathbb{R}$, which induces parameter singularities. Restating this explicitly, it is likely that for some polynomial regression function $f_A$ induced by $A := \{a_i\}_{i = 0}^d \subset \mathbb{R}$ and some polynomial regression function $f_B$ induced by $B := \{b_i\}_{i = 0}^d \subset \mathbb{R}$, the functions $f_A\mid_{\mathcal{X}} = f_B\mid_{\mathcal{X}}$ despite $A \neq B$. Such singularities decrease the LLC \parencite{hoogland2023weirdtrick}, which is consistent with our results. Furthermore, tighter interval choices for $\mathcal{X}$ would lead to \emph{more} singularities and \emph{lower} LLCs -- this is again consistent with our results in Figure \ref{fig:llc_vs_poly_reg}. \emph{This experiment establishes that models on practically constrained domains can often be more singular (and thus have higher generalization capability) than expected.}

\begin{figure}[h!]
\centering
\includegraphics[width=0.7\linewidth]{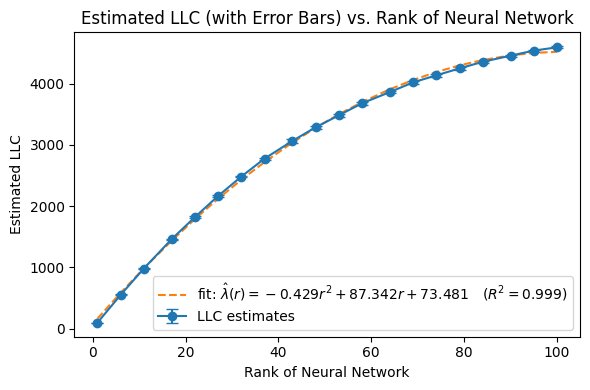}
\caption{Estimated LLC versus rank of neural network, with quadratic fit.}
\label{fig:llc_vs_rank_final}
\end{figure}

\paragraph{Experiment Two: Low-Rank Neural Network with Matrix Factorization.} For our second experiment, we choose a problem that a priori has explicit singularities. Consider a 1-layer neural network $f: \mathbb{R}^d \to \mathbb{R}^d$ given by $f(x) = W_2W_1x$ where $W_2 \in \mathbb{R}^{d \times r}, W_1 \in \mathbb{R}^{r \times d}$ for constants $r \leq d$. Then $\text{rank}(W_2W_1) \leq r$ and moreover this model is singularity-full as for any invertible $G \in \mathbb{R}^{r \times r}$, $W_2W_1 = (W_2G)(G^{-1}W_1)$. We fix $d := 100$ and use the same methodology as in the previous experiment, where we are now testing LLC $\lambda_r$ across 20 choices of rank $r$ linearly spaced from $1$ to $d$. We report our results in Figure \ref{fig:llc_vs_rank_final}.

We can see an extremely strong quadratic fit, peaking when the rank $r = d$. This fit matches the expected results quite nicely. Specifically, it is a standard fact from algebraic geometry that the set $\mathcal{M}_r$ of matrices with rank $r$ is a smooth manifold of dimension $r(2d - r)$. Hence, on that manifold, we can essentially treat our low-rank neural network $f$ as a \emph{regular} model with $r(2d - r)$ parameters $\implies \lambda_r = \frac{1}{2}r(2d - r) \approx -\frac{1}{2}r^2 + 100r$. Note this quadratic matches cleanly with the $-0.429r^2 + 87.342r$ terms in our fit.

\paragraph{Experiment Three: Autoencoders of Increasing Bottleneck Dimension.} For our final experiment, we consider an autoencoder $f: \mathbb{R}^d\to \mathbb{R}^d$. We again fix $d := 100$ and for a given rank $r \leq d$, generate a sample $x \in \mathbb{R}^d$ as $x = Az$ where $z \sim \mathcal{N}(0, I_r)$ and $A \in \mathbb{R}^{d \times r}$. Thus, increasing $r$ increases the dimension of the subspace that our sample $x$ lives in. We design our autoencoder as a sequential concatenation of a ReLU MLP encoder ($d \to 128 \to r$ units) and a ReLU MLP Decoder ($r \to 128 \to d$) units. We train our autoencoders with MSE and present our results in Figure \ref{fig:llc-vs-rank-autoencoder-final}: 

\begin{figure}[h!]
\centering
\includegraphics[width=0.7\linewidth]{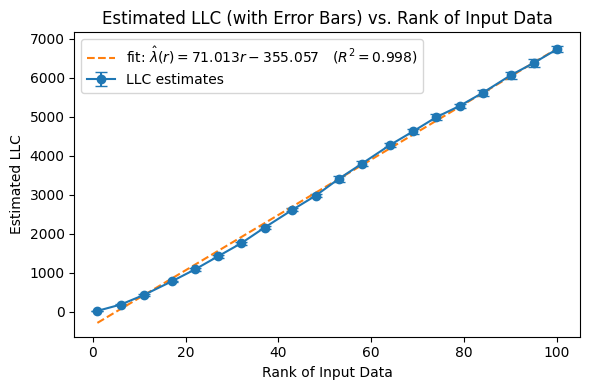}
\caption{Estimated LLC versus rank of input data, with a linear fit.}
\label{fig:llc-vs-rank-autoencoder-final}
\end{figure}

This result is unique because it gives a clean linear fit ($R^2 = 0.998$) despite most MLPs certainly not being regular models (i.e. symmetry-full \& singularity-full.) To the best of our knowledge, we have not seen other scaling law results like this in the SLT literature. Additionally, this result is also meaningful due to the   autoencoder's bottleneck layer effectively performing (non-linear) PCA \parencite{kramer1991nonlinear}. 

\section{Conclusion and Future Directions}

We present a collection of empirical results on using singular learning theory (SLT) as a tool to study both phase transitions and problem complexity in modern neural networks. In Question~1, we obtain mixed evidence for an Arrhenius-style reaction-rate relationship between the change in free energy $\Delta \mathcal{F}_{i \to j}$ and the time-to-transition $r_{i \to j}$ on two popular phase-transition setups (modular addition and Anthropic's Toy Models of Superposition). In Question~2, we show that practically constrained domains can induce singularities in regular models (Experiment~1), and we recover the expected algebraic-geometry prediction $\lambda_r \approx \tfrac{1}{2} r (2d - r)$ for low-rank matrix factorization when $f(x) = W_2 W_1 x$
(Experiment~2). Finally, we demonstrate a peculiar linear fit for the LLC on singular autoencoders of growing bottleneck dimension (Experiment 3).  These results provide concrete examples where SLT reproduces known scaling laws but also sometimes does not align with real-world measurements.

\paragraph{Future Directions.}
We leave the following open questions and directions for future work:

\begin{itemize}
    \item \textbf{Free-energy barriers and reaction rates.}
    Develop a principled method to estimate the free-energy barrier(s) between two subsets of weight space $\mathcal{W}_i$ and $\mathcal{W}_j$. This would enable a more direct test of the Arrhenius-style reaction-rate hypothesis relating $\Delta \mathcal{F}_{i \to j}$
    to $r_{i \to j}$.
    
    \item \textbf{Constrained domains induce singularities.}
    Identify other standard deep-learning settings where natural domain constraints
    (e.g.\ pixel intensities in $[0,255]$) cause distinct parameterizations to yield the same function behavior. Then replicate the polynomial-regressor study of Question~2, Experiment~1 in these settings. This could clarify when such constraints meaningfully lower the LLC, which oftentimes enables generalization.
    
    \item \textbf{Comparing LLC across memorization-vs-generalization architectures.}
    For a fixed task family (e.g.\ image classification), compare the LLC of a heavily over-parameterized ``memorization'' network and a more compact ``generalization'' network. See how this difference scales with problem difficulty (e.g. adding more classification classes)\footnote{We did try this on a sinusoidal regression problem where the memorization and generalization networks were a Wide ReLU MLP and a Linear Regressor in the Fourier Basis respectively. Our results indicated that the MLP was so singularity-full that its LLC was actually \emph{smaller} than that of the generalization network, as would be expected by SLT. While we did not present this result, we posit that future work in this direction could be meaningful.}. 
\end{itemize}

\section{Appendix}

\subsection*{Q1 \& Q2: SGLD Parameters used for LLC Estimation Across All Experiments}

As stated, estimating the LLC relies on Stochastic Gradient Langevin Dynamics (SGLD), which required hyperparameter tuning in order to be numerically stable. For each of our experiments, we provide the learning rate $\epsilon$, choice of $\gamma$, and number of steps $n$ that we used in the below table.
\begin{table}[h]
    \centering
    \label{tab:sgld-params}
    \begin{tabular}{lcccc}
        \hline
        Question \& Experiment \# & $\epsilon$ & $\gamma$ & $n$ \\
        \hline
        Question One, Experiment One & 3e-3  & 5.0 & 500 &  \\
        Question One, Experiment Two & 5e-4 & 1.0 & 400 &  \\
        Question Two, Experiment One & 1e-3 & 1.0 & 2000 &  \\
        Question Two, Experiment Two & 1e-3 & 1.0 & 2000 &  \\
        Question Two, Experiment Three & 1e-5 & 1.0 & 2000 & \\ 
        \hline
    \end{tabular}
    \caption{SGLD hyperparameters used for LLC estimation \parencite{lau2023local} across all experiments.}
\end{table}





\printbibliography

@article{chen2023dynamical,
  title={Dynamical versus bayesian phase transitions in a toy model of superposition},
  author={Chen, Zhongtian and Lau, Edmund and Mendel, Jake and Wei, Susan and Murfet, Daniel},
  journal={arXiv preprint arXiv:2310.06301},
  year={2023}
}

@article{elhage2022toy,
  title   = {Toy Models of Superposition},
  author  = {Elhage, Nelson and Hume, Tristan and Olsson, Catherine and
             Schiefer, Nicholas and Henighan, Tom and Kravec, Shauna and
             Hatfield-Dodds, Zac and Lasenby, Robert and Drain, Dawn and
             Chen, Carol and Grosse, Roger and McCandlish, Sam and
             Kaplan, Jared and Amodei, Dario and Wattenberg, Martin and
             Olah, Christopher},
  journal = {arXiv preprint arXiv:2209.10652},
  year    = {2022},
  url     = {https://arxiv.org/abs/2209.10652}
}

@online{hoogland2023weirdtrick,
  author  = {Jesse Hoogland},
  title   = {Neural Networks Generalize Because of This One Weird Trick},
  year    = {2023},
  month   = jan,
  url     = {https://www.lesswrong.com/posts/fovfuFdpuEwQzJu2w/neural-networks-generalize-because-of-this-one-weird-trick},
  note    = {LessWrong}
}

@article{arrhenius1889reaktionsgeschwindigkeit,
  title={{\"U}ber die Reaktionsgeschwindigkeit bei der Inversion von Rohrzucker durch S{\"a}uren},
  author={Arrhenius, Svante},
  journal={Zeitschrift f{\"u}r physikalische Chemie},
  volume={4},
  number={1},
  pages={226--248},
  year={1889},
  publisher={De Gruyter (O)}
}

@article{kramer1991nonlinear,
  title   = {Nonlinear Principal Component Analysis Using Autoassociative Neural Networks},
  author  = {Kramer, Mark A.},
  journal = {AIChE Journal},
  volume  = {37},
  number  = {2},
  pages   = {233--243},
  year    = {1991},
  doi     = {10.1002/aic.690370209}
}

@article{watanabe2010equations,
  title={Equations of states in singular statistical estimation},
  author={Watanabe, Sumio},
  journal={Neural Networks},
  volume={23},
  number={1},
  pages={20--34},
  year={2010},
  publisher={Elsevier}
}

@article{power2022grokking,
  title   = {Grokking: Generalization Beyond Overfitting on Small Algorithmic Datasets},
  author  = {Power, Alethea and Burda, Yuri and Edwards, Harri and Babuschkin, Igor and Misra, Vedant},
  journal = {arXiv preprint arXiv:2201.02177},
  year    = {2022}
}

@article{lehalleur2025you,
  title={You Are What You Eat--AI Alignment Requires Understanding How Data Shapes Structure and Generalisation},
  author={Lehalleur, Simon Pepin and Hoogland, Jesse and Farrugia-Roberts, Matthew and Wei, Susan and Oldenziel, Alexander Gietelink and Wang, George and Carroll, Liam and Murfet, Daniel},
  journal={arXiv preprint arXiv:2502.05475},
  year={2025}
}

@misc{pearce2023memorize,
  title        = {Do Machine Learning Models Memorize or Generalize?},
  author       = {Pearce, Adam and Ghandeharioun, Asma and Hussein, Nada and Thain, Nithum and Wattenberg, Martin and Dixon, Lucas},
  howpublished = {\url{https://pair.withgoogle.com/explorables/grokking/}},
  note         = {Explorables, PAIR (People + AI Research) at Google},
  month        = aug,
  year         = {2023},
  urldate      = {2025-10-27}
}

@book{watanabe2009,
  author    = {Sumio Watanabe},
  title     = {Algebraic Geometry and Statistical Learning Theory},
  year      = {2009},
  publisher = {Cambridge University Press}
}

@book{watanabe2018,
  author    = {Sumio Watanabe},
  title     = {Mathematical Theory of Bayesian Statistics},
  year      = {2018},
  publisher = {Cambridge University Press}
}

@article{lau2023local,
  title={The local learning coefficient: A singularity-aware complexity measure},
  author={Lau, Edmund and Furman, Zach and Wang, George and Murfet, Daniel and Wei, Susan},
  journal={arXiv preprint arXiv:2308.12108},
  year={2023}
}

@article{wei2022deep,
  title={Deep learning is singular, and that’s good},
  author={Wei, Susan and Murfet, Daniel and Gong, Mingming and Li, Hui and Gell-Redman, Jesse and Quella, Thomas},
  journal={IEEE Transactions on Neural Networks and Learning Systems},
  volume={34},
  number={12},
  pages={10473--10486},
  year={2022},
  publisher={IEEE}
}

@online{panickssery2023learningcoef,
  author       = {Nina Panickssery and Dmitry Vaintrob},
  title        = {Investigating the learning coefficient of modular addition: hackathon project},
  date         = {2023-10-17},
  url          = {https://www.lesswrong.com/posts/4v3hMuKfsGatLXPgt/investigating-the-learning-coefficient-of-modular-addition},
  urldate      = {2025-09-22},
  organization = {LessWrong}
}

\end{document}